%% file: samplepaper.tex
\documentclass[runningheads]{llncs}

\usepackage{graphicx}
\usepackage{amsmath,amssymb} % define this before the line numbering.
\usepackage{bm}
\usepackage{color}
\usepackage{algorithmicx}
\usepackage{algorithm}
\usepackage{algpseudocode}
\usepackage{subfigure}
\usepackage{xcolor}
\usepackage{multirow}
\usepackage{booktabs}
\usepackage{sidecap}

\newcommand{\myparagraph}[1]{{\noindent \bf #1}}

\usepackage{nicefrac}       % compact symbols for 1/2, etc.
\usepackage{xfrac}
\usepackage{microtype}      % microtypography
\usepackage{mathtools}
\usepackage{changepage}

\usepackage{xspace}
\usepackage{wrapfig}

\usepackage{color}
\usepackage{multirow}

\usepackage{graphicx}
\usepackage{anyfontsize}
\usepackage{caption}

\graphicspath{{./images/}}

\makeatletter
\DeclareRobustCommand\onedot{\futurelet\@let@token\@onedot}
\def\@onedot{\ifx\@let@token.\else.\null\fi\xspace}

\def\eg{\emph{e.g}\onedot} 
\def\ie{\emph{i.e}\onedot}

\def\wrt{w.r.t\onedot} 

\makeatother

\usepackage{authblk}
\usepackage[symbol*]{footmisc}

\DefineFNsymbolsTM{otherfnsymbols}{%
  \textasteriskcentered *
  \textdagger    \dagger
  \textdaggerdbl \ddagger
  \textsection   \mathsection
  \textasteriskcentered *
  \textbardbl    \|%
  \textparagraph \mathparagraph
}%

\setfnsymbol{otherfnsymbols}

\begin{document}

\title{Acquisition of Localization Confidence for Accurate Object Detection}
\titlerunning{Acquisition of Localization Confidence for Accurate Object Detection}
\authorrunning{B. Jiang, R. Luo, J. Mao, T. Xiao, and Y. Jiang}
\author{
  ~~~~~~~~~Borui Jiang\thanks{~indicates equal contribution.}\inst{1,3} \and
  Ruixuan Luo${}^*$\inst{1,3} \and
  Jiayuan Mao${}^*$\inst{2,4} \and \newline
  Tete Xiao\inst{1,3} \and
  Yuning Jiang\inst{4}
}
\institute{
School of Electronics Engineering and Computer Science, Peking University \and
ITCS, Institute for Interdisciplinary Information Sciences, Tsinghua University \and
Megvii Inc. (Face++) \qquad ${}^4$~Toutiao AI Lab\\
\email{\{jbr, luoruixuan97, jasonhsiao97\}@pku.edu.cn,\\
mjy14@mails.tsinghua.edu.cn, jiangyuning@bytedance.com}}

\maketitle
\begin{abstract}
Modern CNN-based object detectors rely on bounding box regression and non-maximum suppression to localize objects. While the probabilities for class labels naturally reflect classification confidence, localization confidence is absent. This makes properly localized bounding boxes degenerate during iterative regression or even suppressed during NMS. In the paper we propose IoU-Net learning to predict the IoU between each detected bounding box and the matched ground-truth. The network acquires this confidence of localization, which improves the NMS procedure by preserving accurately localized bounding boxes. Furthermore, an optimization-based bounding box refinement method is proposed, where the predicted IoU is formulated as the objective. Extensive experiments on the MS-COCO dataset show the effectiveness of IoU-Net, as well as its compatibility with and adaptivity to several state-of-the-art object detectors.
\end{abstract}

\input{src/introduction}
\input{src/delving}
\input{src/nms}
\input{src/refinement}
\input{src/joint}
\input{src/experiments}

% *** Section 7:  Conclusion *** %
\section{Conclusion}
In this paper, a novel network architecture, namely IoU-Net, is proposed for accurate object localization. By learning to predict the IoU with matched ground-truth, IoU-Net acquires ``localization confidence'' for the detected bounding box. This empowers an IoU-guided NMS procedure where accurately localized bounding boxes are prevented from being suppressed.
The proposed IoU-Net is intuitive and can be easily integrated into a broad set of detection models to improve their localization accuracy. Experimental results on MS-COCO demonstrate its effectiveness and potential in practical applications.

This paper points out the misalignment of classification and localization confidences in modern detection pipelines.
We also formulate an novel optimization view on the problem of bounding box refinement, and the proposed solution surpasses the regression-based methods.
We hope these novel viewpoints provide insights to future works on object detection, and beyond.

\bibliographystyle{splncs04}
\bibliography{egbib}

\end{document}

%% file: src/introduction.tex
\section{Introduction}
Object detection serves as a prerequisite for a broad set of downstream vision applications, such as instance segmentation \cite{pinheiro2015learning,pinheiro2016learning}, human skeleton \cite{toshev2014deeppose}, face recognition \cite{taigman2014deepface} and high-level object-based reasoning \cite{wu2017learning}.
Object detection combines both object classification and object localization.
A majority of modern object detectors are based on two-stage frameworks~\cite{Girshick_2014_CVPR,Girshick_2015_ICCV,ren2015faster,Lin_2017_CVPR,he2017mask}, in which object detection is formulated as a multi-task learning problem: 1) distinguish foreground object proposals from background and assign them with proper class labels; 2) regress a set of coefficients which localize the object by maximizing intersection-over-union (IoU) or other metrics between detection results and the ground-truth. Finally, redundant bounding boxes (duplicated detections on the same object) are removed by a non-maximum suppression (NMS) procedure.

Classification and localization are solved differently in such detection pipeline. Specifically, given a proposal, while the probability for each class label naturally acts as an ``classification confidence'' of the proposal, the bounding box regression module finds the optimal transformation for the proposal to best fit the ground-truth. However, the ``localization confidence'' is absent in the loop.

\begin{figure}[!t]
  % class score & real iou
  \centering
  \subfigure[Demonstrative cases of the misalignment between classification confidence and localization accuracy. The yellow bounding boxes denote the ground-truth, while the red and green bounding boxes are both detection results yielded by FPN~\cite{Lin_2017_CVPR}. Localization confidence is computed by the proposed IoU-Net. Using classification confidence as the ranking metric will cause accurately localized bounding boxes (in green) being incorrectly eliminated in the traditional NMS procedure. Quantitative analysis is provided in Section~\ref{sec:problem:nms}]{
     \includegraphics[width=\textwidth]{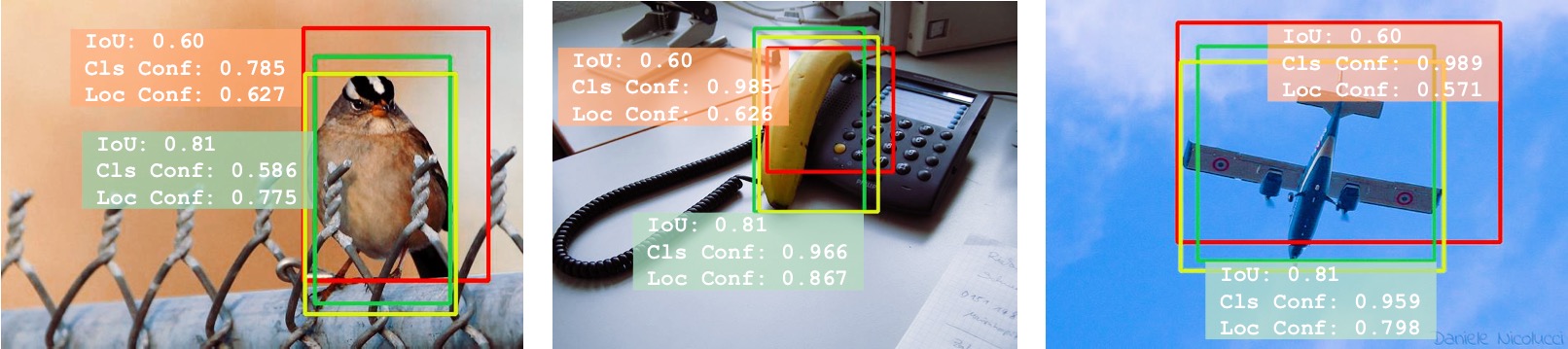}
    \label{fig:example:a}
  }
  \subfigure[Demonstrative cases of the non-monotonic localization in iterative bounding box regression. Quantitative analysis is provided in Section~\ref{sec:problem:bbreg}.]{
     \includegraphics[width=\textwidth,trim={0.4cm 0 0.2cm 0}]{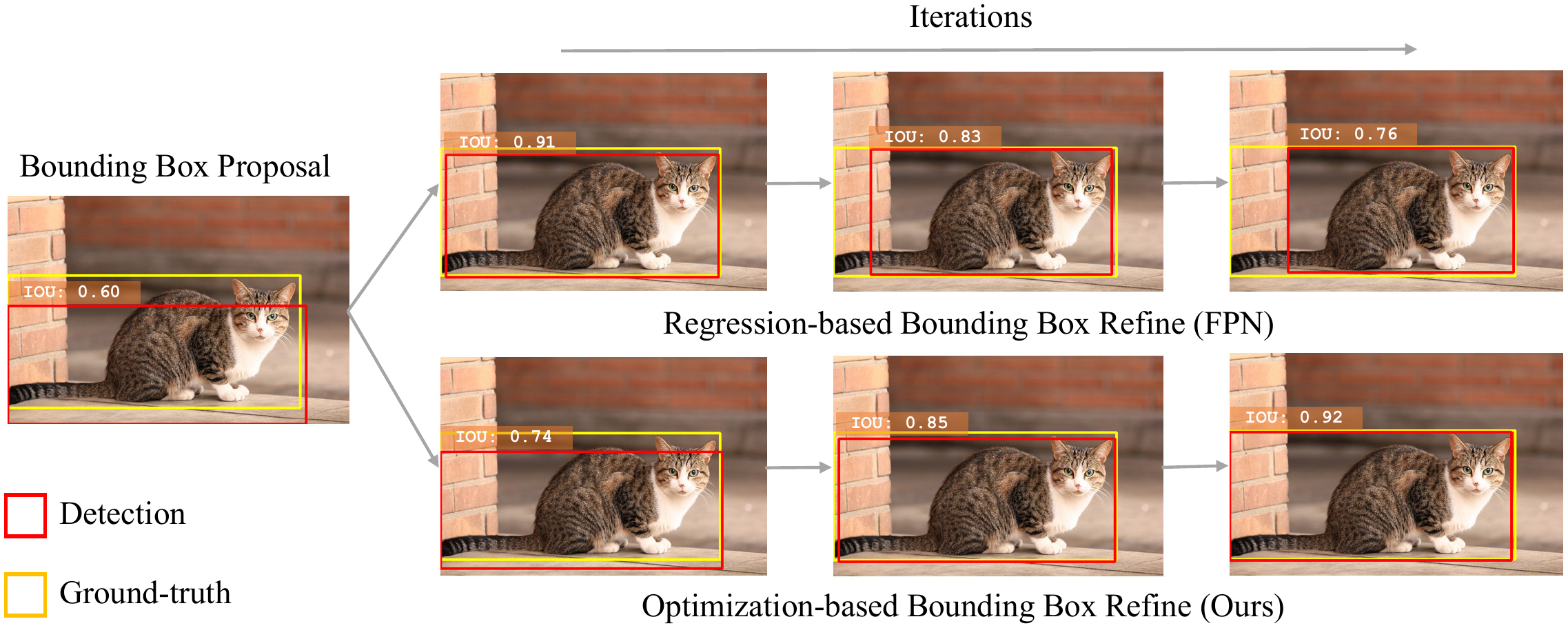}
    \label{fig:example:b}
  }
  \caption{Visualization on two drawbacks brought by the absence of localization confidence. Examples are selected from MS-COCO \emph{minival} \cite{lin2014microsoft}.}%\newline
%\textbf{Top~row:} Demonstration of the misalignment between classification confidence and localization accuracy. The yellow bounding box denote the ground-truth box, while the red bounding boxes is the proposals generated by FPN \cite{Lin_2017_CVPR}. Localization confidence are computed by IoU-Net. In the traditional NMS procedure, bounding boxes with more accurate localization with be incorrectly eliminated. Using localization confidence alleviates this problem. Quantitative analysis is provided in Section~\ref{sec:problem:nms}.\newline
%\textbf{Bottom row:} Demonstration of the non-monotonic localization in iterative bounding box regression. The proposed optimization-based bounding box refinement provides monotonic and robust improvement. Quantitative analysis is provided in Section~\ref{sec:problem:bbreg}.}
  \label{fig:example}
%   \vspace{-3em}
\end{figure}

This brings about two drawbacks.
(1) First, the suppression of duplicated detections is ignorant of the localization accuracy while the classification scores are typically used as the metric for ranking the proposals. In Figure~\ref{fig:example:a}, we show a set of cases where the detected bounding boxes with higher classification confidences contrarily have smaller overlaps with the corresponding ground-truth. Analog to Gresham's saying that {\it bad money drives out good}, the misalignment between classification confidence and localization accuracy may lead to accurately localized bounding boxes being suppressed by less accurate ones in the NMS procedure. (2) Second, the absence of localization confidence makes the widely-adopted bounding box regression less interpretable. As an example, previous works~\cite{cai2017cascade} report the non-monotonicity of iterative bounding box regression. That is, bounding box regression may degenerate the localization of input bounding boxes if applied for multiple times (shown as Figure~\ref{fig:example:b}).

In this paper we introduce IoU-Net, which predicts the IoU between detected bounding boxes and their corresponding ground-truth boxes, making the networks aware of the localization criterion analog to the classification module. This simple coefficient provides us with new solutions to the aforementioned problems:

\begin{enumerate}
    \item IoU is a natural criterion for localization accuracy. We can replace classification confidence with the predicted IoU as the ranking keyword in NMS. This technique, namely IoU-guided NMS, help to eliminate the suppression failure caused by the misleading classification confidences.
    \item We present an optimization-based bounding box refinement procedure on par with the traditional regression-based methods. During the inference, the predicted IoU is used as the optimization objective, as well as an interpretable indicator of the localization confidence. The proposed Precise RoI Pooling layer enables us to solve the IoU optimization by gradient ascent. We show that compared with the regression-based method, the optimization-based bounding box refinement empirically provides a monotonic improvement on the localization accuracy. The method is fully compatible with and can be integrated into various CNN-based detectors \cite{Lin_2017_CVPR,cai2017cascade,he2017mask}.
\end{enumerate}

%% file: src/delving.tex
\section{Delving into object localization}
\label{sec:delving}

First of all, we explore two drawbacks in object localization: the misalignment between classification confidence and localization accuracy and the non-monotonic bounding box regression. A standard FPN~\cite{Lin_2017_CVPR} detector is trained on MS-COCO \emph{trainval35k} as the baseline and tested on \emph{minival} for the study.

\subsection{Misaligned classification and localization accuracy}
\label{sec:problem:nms}
\begin{figure}[t]
\centering
\subfigure[IoU vs. Classification Confidence]{
    \includegraphics[width=0.47\columnwidth]{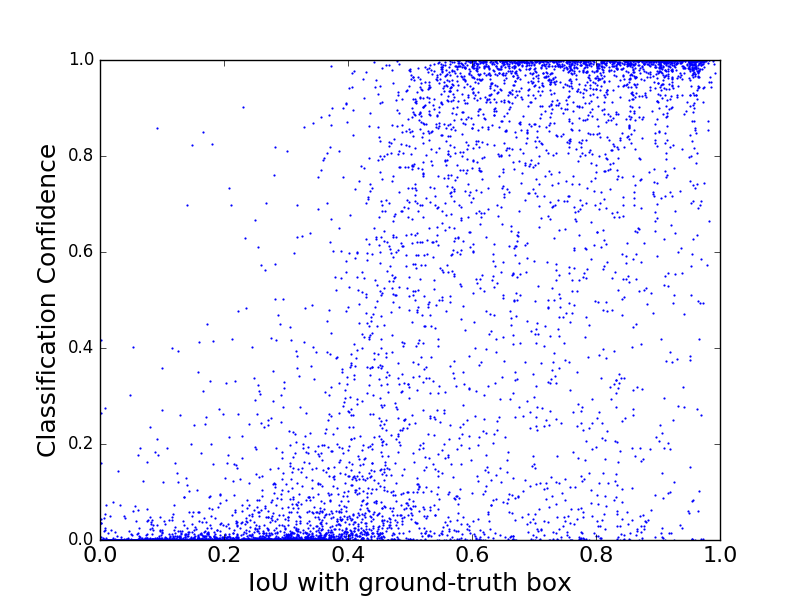}
    \label{fig:distribution:cls}
}
\subfigure[IoU vs. Localization Confidence]{
    \includegraphics[width=0.47\columnwidth]{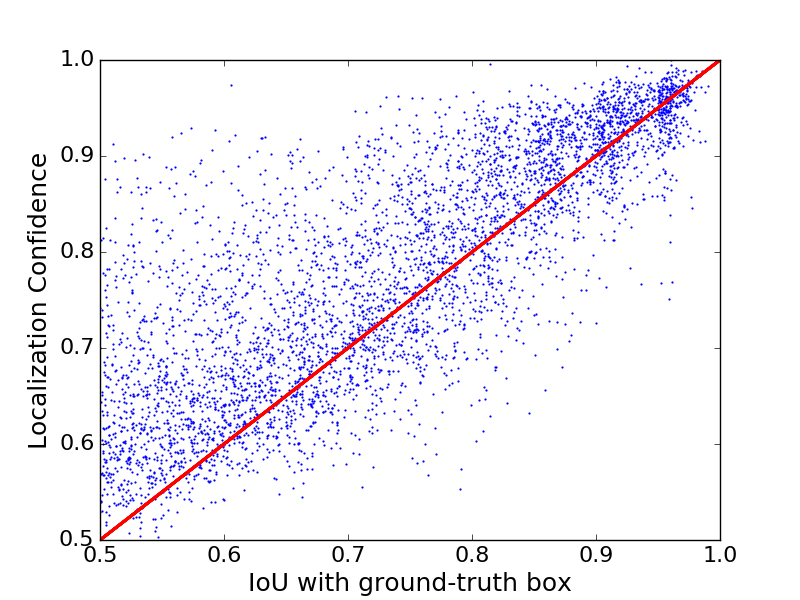}
    \label{fig:distribution:iou}
}
\caption{
The correlation between the IoU of bounding boxes with the matched ground-truth and the classification/localization confidence. Considering detected bounding boxes having an IoU ($> 0.5$) with the corresponding ground-truth, the Pearson correlation coefficients are: (a) 0.217, and (b) 0.617.\newline
(a) The classification confidence indicates the category of a bounding box, but cannot be interpreted as the localization accuracy.\newline
(b) To resolve the issue, we propose IoU-Net to predict the localization confidence for each detected bounding box, \ie, its IoU with corresponding ground-truth.
}
\label{fig:distribution}
\end{figure}

With the objective to remove duplicated bounding boxes, NMS has been an indispensable component in most object detectors since~\cite{dalal2005histograms}. NMS works in an iterative manner. At each iteration, the bounding box with the maximum classification confidence is selected and its neighboring boxes are eliminated using a predefined overlapping threshold. In Soft-NMS~\cite{bodla2017improving} algorithm, box elimination is replaced by the decrement of confidence, leading to a higher recall. Recently, a set of learning-based algorithms have been proposed as alternatives to the parameter-free NMS and Soft-NMS. \cite{rothe2014non} calculates an overlap matrix of all bounding boxes and performs affinity propagation clustering to select exemplars of clusters as the final detection results. \cite{hosang2017learning} proposes the GossipNet, a post-processing network trained for NMS based on bounding boxes and the classification confidence. \cite{hu2017relation} proposes an end-to-end network learning the relation between detected bounding boxes. However, these parameter-based methods require more computational resources which limits their real-world application.

In the widely-adopted NMS approach, the classification confidence is used for ranking bounding boxes, which can be problematic.
We visualize the distribution of classification confidences of all detected bounding boxes before NMS, as shown in Figure~\ref{fig:distribution:cls}. The x-axis is the IoU between the detected box and its matched ground-truth, while the y-axis denotes its classification confidence. The Pearson correlation coefficient indicates that the localization accuracy is not well correlated with the classification confidence.
% , indicating a misalignment between classification confidence and localization accuracy.

We attribute this to the objective used by most of the CNN-based object detectors in distinguishing foreground (positive) samples from background (negative) samples. A detected bounding box $box_{\text{det}}$ is considered positive during training if its IoU with one of the ground-truth bounding box is greater than a threshold $\Omega_{train}$.
This objective can be misaligned with the localization accuracy. 
Figure~\ref{fig:example:a} shows cases where bounding boxes having higher classification confidence have poorer localization. 

\begin{SCfigure}
\centering
\includegraphics[width=0.5\columnwidth,trim={0 0 0.6cm 0}]{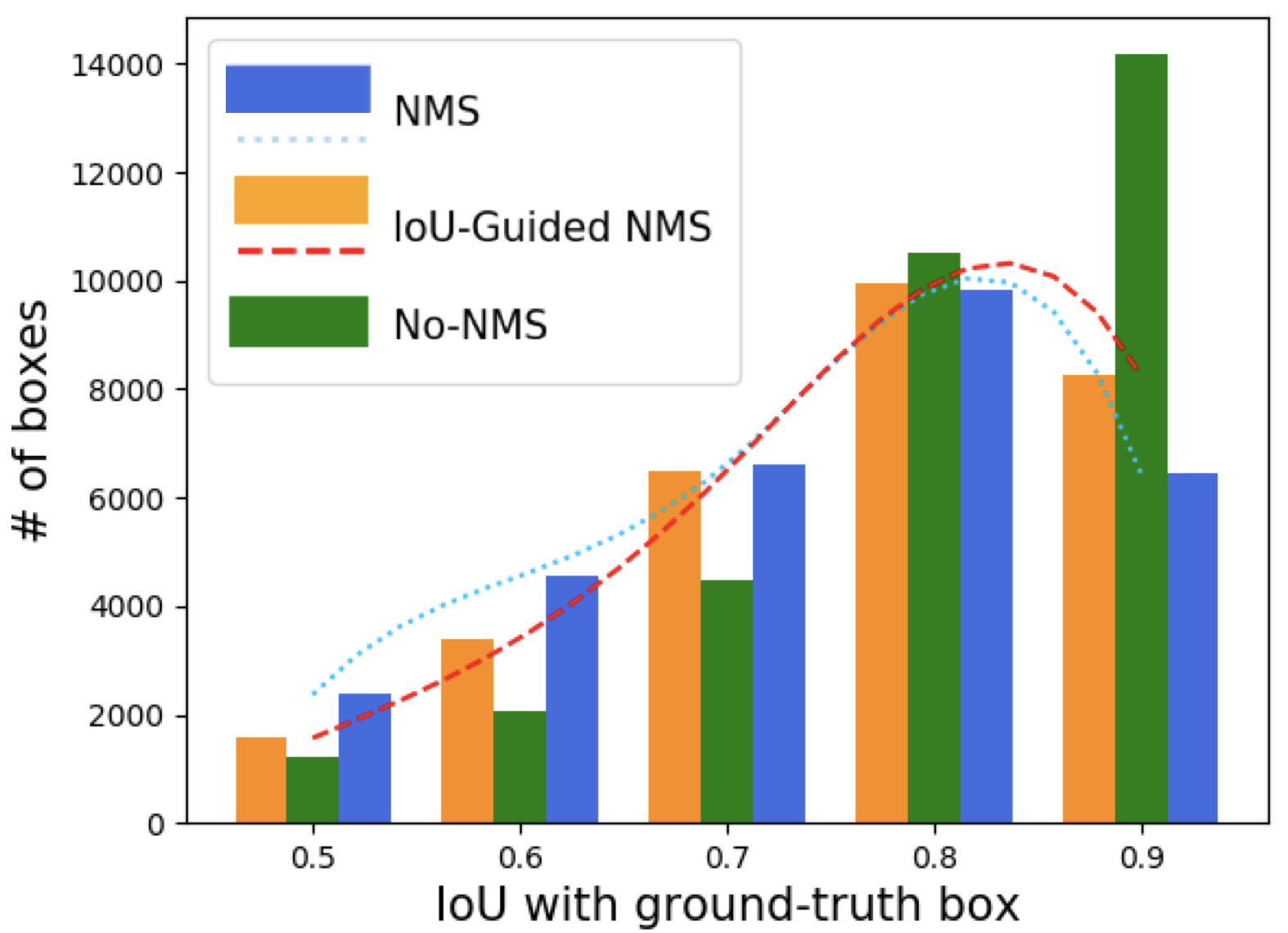}
\caption{
The number of positive bounding boxes after the NMS, grouped by their IoU with the matched ground-truth. 
In traditional NMS (blue bar), a significant portion of accurately localized bounding boxes get mistakenly suppressed due to the misalignment of classification confidence and localization accuracy, while IoU-guided NMS (yellow bar) preserves more accurately localized bounding boxes.
}
\label{fig:ioufreq:after}
\end{SCfigure}

Recall that in traditional NMS, when there exists duplicated detections for a single object, the bounding box with maximum classification confidence will be preserved. However, due to the misalignment, the bounding box with better localization will probably get suppressed during the NMS, leading to the poor localization of objects. Figure~\ref{fig:ioufreq:after} quantitatively shows the number of positive bounding boxes after NMS. The bounding boxes are grouped by their IoU with the matched ground-truth. For multiple detections matched with the same ground-truth, only the one with the highest IoU is considered positive. Therefore, No-NMS could be considered as the upper-bound for the number of positive bounding boxes. We can see that the absence of localization confidence makes more than half of detected bounding boxes with IoU $> 0.9$ being suppressed in the traditional NMS procedure, which degrades the localization quality of the detection results.

%%%%%%%%%%%%%%%%%%%%%%%%%%%%%%%%%%%%%%%%%%%%%%%%%%%%%%%%%%%%%%%%%%%%%%%%%%
%%%%%%%%%%%%%%%%%%%%%%%%%%%%%%%%%%%%%%%%%%%%%%%%%%%%%%%%%%%%%%%%%%%%%%%%%%
%%%%%%%%%%%%%%%%%%%%%%%%%%%%%%%%%%%%%%%%%%%%%%%%%%%%%%%%%%%%%%%%%%%%%%%%%%
%%%%%%%%%%%%%%%%%%%%%%%%%%%%%%%%%%%%%%%%%%%%%%%%%%%%%%%%%%%%%%%%%%%%%%%%%%
%%%%%%%%%%%%%%%%%%%%%%%%%%%%%%%%%%%%%%%%%%%%%%%%%%%%%%%%%%%%%%%%%%%%%%%%%%

\subsection{Non-monotonic bounding box regression}
\label{sec:problem:bbreg}

\begin{figure}[t]
\centering
\subfigure[FPN]{ 
    \includegraphics[width=0.45\columnwidth]{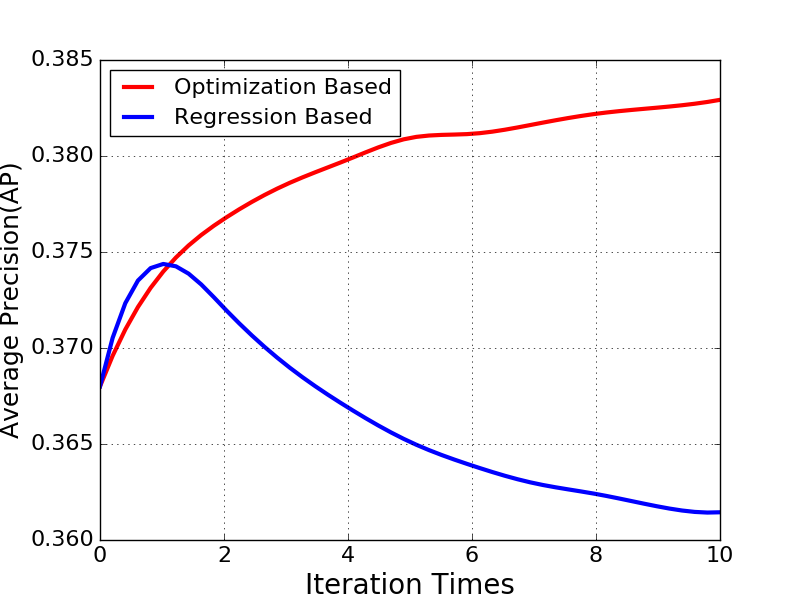}
    \label{fig:apiter:refine}
}\quad
\subfigure[Cascade R-CNN]{
    \includegraphics[width=0.45\columnwidth]{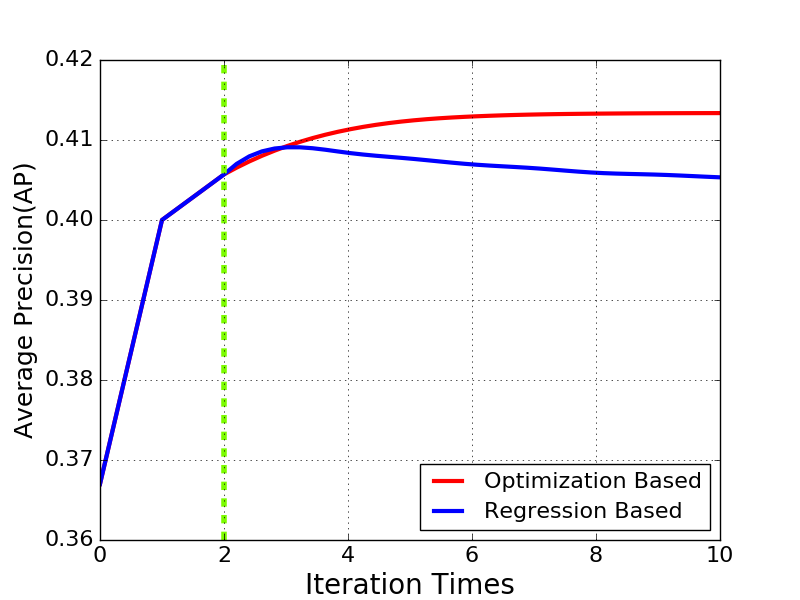}
    \label{fig:apiter:cascade}
}
\caption{Optimization-based \emph{v.s.} Regression-based BBox refinement. \textbf{(a)} Comparison in FPN. When applying the regression iteratively, the AP of detection results firstly get improved but drops quickly in later iterations. \textbf{(b)} Camparison in Cascade R-CNN. Iteration 0, 1 and 2 represents the 1st, 2nd and 3rd regression stages in Cascade R-CNN. For iteration $i \ge 3$, we refine the bounding boxes with the regressor of the third stage. After multiple iteration, AP slightly drops, while the optimization-based method further improves the AP by 0.8\%.
}
\label{fig:apiter}
\end{figure}

In general, single object localization can be classified into two categories: bounding box-based methods and segment-based methods. The segment-based methods~\cite{pinheiro2015learning,pinheiro2016learning,hu2017fastmask,he2017mask} aim to generate a pixel-level segment for each instance but inevitably require additional segmentation annotation. This work focuses on the bounding box-based methods.

%Sliding-window strategy is widely adopted in the earliest object detectors~\cite{viola2001rapid,dollar2014fast,nam2014local} to locate objects in images. It could only output detected bounding boxes in discrete steps of aspect ratios and scales.
Single object localization is usually formulated as a bounding box regression task. The core idea is that a network directly learns to transform (\ie, scale or shift) a bounding box to its designated target.
In~\cite{Girshick_2014_CVPR,Girshick_2015_ICCV} linear regression or fully-connected layer is applied to refine the localization of object proposals generated by external pre-processing modules (\eg, Selective Search~\cite{uijlings2013selective} or EdgeBoxes~\cite{zitnick2014edge}). Faster R-CNN~\cite{NIPS2015_5638} proposes region proposal network (RPN) in which only predefined anchors are used to train an end-to-end object detector. \cite{huang2015densebox,yu2016unitbox} utilize anchor-free, fully-convolutional networks to handle object scale variation. Meanwhile, Repulsion Loss is proposed in~\cite{wang2017repulsion} to robustly detect objects with crowd occlusion.
Due to its effectiveness and simplicity, bounding box regression has become an essential component in most CNN-based detectors. 

A broad set of downstream applications such as tracking and recognition will benefit from accurately localized bounding boxes. This raises the demand for improving localization accuracy. In a series of object detectors~\cite{yang2016craft,gidaris2016attend,gidaris2015object,rajaram2016refinenet}, refined boxes will be fed to the bounding box regressor again and go through the refinement for another time. This procedure is performed for several times, namely iterative bounding box regression. Faster R-CNN~\cite{NIPS2015_5638} first performs the bounding box regression twice to transform predefined anchors into final detected bounding boxes. \cite{li2017multi} proposes a group recursive learning approach to iteratively refine detection results and minimize the offsets between object proposals and the ground-truth considering the global dependency among multiple proposals. G-CNN is proposed in~\cite{najibi2016g} which starts with a multi-scale regular grid over the image and iteratively pushes the boxes in the grid towards the ground-truth.
However, as reported in~\cite{cai2017cascade}, applying bounding box regression more than twice brings no further improvement. \cite{cai2017cascade} attribute this to the distribution mismatch in multi-step bounding box regression and address it by a resampling strategy in multi-stage bounding box regression.

We experimentally show the performance of iterative bounding box regression based on FPN and Cascade R-CNN frameworks. The Average Precision (AP) of the results after each iteration are shown as the blue curves in Figure~\ref{fig:apiter:refine} and Figure~\ref{fig:apiter:cascade}, respectively.
The AP curves in Figure~\ref{fig:apiter} show that the improvement on localization accuracy, as the number of iterations increase, is non-monotonic for iterative bounding box regression. The non-monotonicity, together with the non-interpretability, brings difficulties in applications.
% As analyzed in~\cite{cai2017cascade}, it is partially because the regression-based method is easy to overfit the distribution of training bounding boxes. 
Besides, without localization confidence for detected bounding boxes, we can not have fine-grained control over the refinement, such as using an adaptive number of iterations for different bounding boxes.

%% file: src/nms.tex
\section{IoU-Net}
To quantitatively analyze the effectiveness of IoU prediction, we first present the methodology adopted for training an IoU predictor in Section~\ref{sec:predictiou}. In Section~\ref{sec:iouguided NMS} and Section~\ref{sec:refinement}, we show how to use IoU predictor for NMS and bounding box refinement, respectively. Finally in Section~\ref{sec:joint} we integrate the IoU predictor into existing object detectors such as FPN \cite{Lin_2017_CVPR}.

\subsection{Learning to predict IoU}
\label{sec:predictiou}
\begin{figure}[!t]
    \centering
    \includegraphics[width=\textwidth]{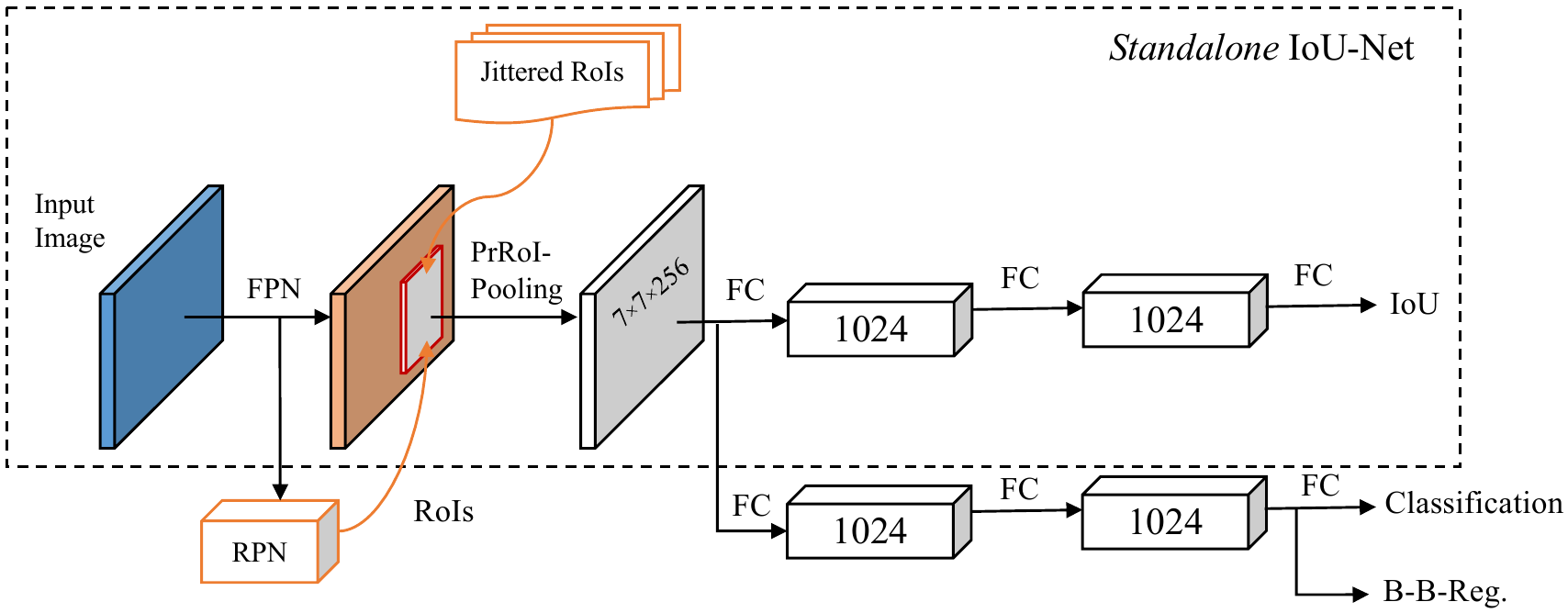}
    \caption{Full architecture of the proposed IoU-Net described in Section~\ref{sec:joint}. Input images are first fed into an FPN backbone. The IoU predictor takes the output features from the FPN backbone. We replace the RoI Pooling layer with a PrRoI Pooling layer described in Section~\ref{sec:refinement}. The IoU predictor shares a similar structure with the R-CNN branch. The modules marked within the dashed box form a \emph{standalone} IoU-Net.
%Since the imbalance of positive and negative samples generated by the RPN framework, our training methodology has a little difference from \ref{}. 
}
    \label{fig:network}
\end{figure}

Shown in Figure~\ref{fig:network}, the IoU predictor takes visual features from the FPN and estimates the localization accuracy (IoU) for each bounding box.
We generate bounding boxes and labels for training the IoU-Net by augmenting the ground-truth, instead of taking proposals from RPNs. Specifically, for all ground-truth bounding boxes in the training set, we manually transform them with a set of randomized parameters, resulting in a candidate bounding box set. We then remove from this candidate set the bounding boxes having an IoU less than $\Omega_{train} = 0.5$ with the matched ground-truth. We uniformly sample training data from this candidate set \wrt the IoU. This data generation process empirically brings better performance and robustness to the IoU-Net. For each bounding box, the features are extracted from the output of FPN with the proposed Precise RoI Pooling layer (see Section~\ref{sec:refinement}). The features are then fed into a two-layer feedforward network for the IoU prediction. For a better performance, we use class-aware IoU predictors.

The IoU predictor is compatible with most existing RoI-based detectors. 
The accuracy of a \emph{standalone} IoU predictor can be found in Figure~\ref{fig:distribution}. As the training procedure is independent of specific detectors, it is robust to the change of the input distributions (\eg, when cooperates with different detectors).
In later sections, we will further demonstrate how this module can be jointly optimized in a full detection pipeline (\ie, jointly with RPNs and R-CNN).

\subsection{IoU-guided NMS}
\label{sec:iouguided NMS}
We resolve the misalignment between classification confidence and localization accuracy with a novel IoU-guided NMS procedure, where the classification confidence and localization confidence (an estimation of the IoU) are  disentangled.
In short, we use the predicted IoU instead of the classification confidence as the ranking keyword for bounding boxes. Analog to the traditional NMS, the box having the highest IoU with a ground-truth will be selected to eliminate all other boxes having an overlap greater than a given threshold $\Omega_{\text{nms}}$. To determine the classification scores, when a box $i$ eliminates box $j$, we update the classification confidence $s_i$ of box $i$ by $s_i = \max(s_i, s_j)$. This procedure can also be interpreted as a confidence clustering: for a group of bounding boxes matching the same ground-truth, we take the most confident prediction for the class label. A psuedo-code for this algorithm can be found in Algorithm~\ref{alg:iounms}.

\begin{algorithm}[!t]
\caption{IoU-guided NMS. Classification confidence and localization confidence are disentangled in the algorithm. We use the localization confidence (the predicted IoU) to rank all detected bounding boxes, and update the classification confidence based on a clustering-like rule.}
\label{alg:iounms}
\begin{algorithmic}[1]
\Require{$ \mathcal{B}=\{b_1,...,b_n\},~\mathcal{S},~\mathcal{I},~\Omega_{\text{nms}}$ \newline $\mathcal{B}$ is a set of detected bounding boxes. \newline $\mathcal{S}$ and $\mathcal{I}$ are functions (neural networks) mapping bounding boxes to their classification confidence and IoU estimation (localization confidence) respectively. \newline $\Omega_{\text{nms}}$ is the NMS threshold.}
\Ensure{$\mathcal{D}$, the set of detected bounding boxes with classification scores.}
\State{$ \mathcal{D} \leftarrow{} \varnothing $}
\While{$ \mathcal{B} \neq \varnothing $}
    \State{$b_m \leftarrow{} \arg\max{\mathcal{I}(b_j)} $}
    \State{$\mathcal B \leftarrow{} \mathcal B \setminus \{b_m\}$}
    \State{$s \leftarrow \mathcal S(b_m)$}
    \For{$b_j \in \mathcal{B}$}
        \If{$\mathrm{IoU(b_m,b_j)>\Omega_{\text{nms}}}$}
            \State{$s \leftarrow{max(s, \mathcal S(b_j))}$}
            \State{$\mathcal B \leftarrow{} \mathcal B \setminus \{b_j\}$}
        \EndIf
    \EndFor
    \State{$\mathcal{D} \leftarrow{\mathcal{D}\cup \{\langle b_m, s \rangle\}}$}
\EndWhile
\State{\Return{$\mathcal{D}$}}
\end{algorithmic}
\end{algorithm}

IoU-guided NMS resolves the misalignment between classification confidence and localization accuracy.
% Shown in Figure~\ref{fig:example}, IoU-guided NMS will keep the detection with better localization accuracy. 
Quantitative results show that our method outperforms traditional NMS and other variants such as Soft-NMS \cite{bodla2017improving}. Using IoU-guided NMS as the post-processor further pushes forward the performance of several state-of-the-art object detectors.

%% file: src/refinement.tex
\subsection{Bounding box refinement as an optimization procedure}
\label{sec:refinement}

\begin{algorithm}[!t]
\caption{Optimization-based bounding box refinement}
\label{alg:bbrefine}
\begin{algorithmic}[1]
\Require{$ \mathcal{B}=\{b_1,...,b_n\},~\mathcal F,~T,~\lambda, ~\Omega_1,~\Omega_2$ \newline $\mathcal{B}$ is a set of detected bounding boxes, in the form of $(x_0, y_0, x_1, y_1)$. \newline $\mathcal{F}$ is the feature map of the input image.
\newline $T$ is number of steps. $\lambda$ is the step size, and $\Omega_1$ is an early-stop threshold and $\Omega_2 < 0$ is an localization degeneration tolerance.\newline
Function $\mathrm{PrPool}$ extracts the feature representation for a given bounding box and function $\mathrm{IoU}$ denotes the estimation of IoU by the IoU-Net.}
\Ensure{The set of final detection bounding boxes.}
\State{$\mathcal{A} \gets \varnothing$}
\For{$i = 1~to~T$}
    \For{$b_j\in\mathcal{B}$ and $b_j\notin\mathcal{A}$}
        \State{$\bm{grad} \leftarrow \nabla_{bj} \mathrm{IoU(PrPool}(\mathcal{F}, b_j))$}
        \State{$PrevScore\leftarrow{\mathrm{IoU(PrPool}(\mathcal{F}, b_j))}$}
        \State{$b_j \leftarrow b_j + \lambda * scale(\bm{grad}, b_j)$}
        \State{$NewScore\leftarrow{\mathrm{IoU(PrPool}(\mathcal{F}, b_j))}$}
        \If{$|PrevScore-NewScore| < \Omega_1$ \textbf{or} $NewScore - PrevScore < \Omega_2$}
            \State{$\mathcal{A} \gets \mathcal A \cup \{b_j\}$}
        \EndIf
    \EndFor
\EndFor
\State{\Return{$\mathcal{B}$}}
\end{algorithmic}
\end{algorithm}

The problem of bounding box refinement can formulated mathematically as finding the optimal $c^*$ \emph{s$.$t$.$}:
\begin{equation}
c^* = \arg \min_c crit {\large(} \textbf{transform}(box_{\text{det}}, c), box_{\text{gt}} \large{)}\,,
\label{eq:optimization}
\end{equation}
where $box_{\text{det}}$ is the detected bounding box, $box_{\text{gt}}$ is a (targeting) ground-truth bounding box and $\textbf{transform}$ is a bounding box transformation function taking $c$ as parameter and transform the given bounding box. $crit$ is a criterion measuring the distance between two bounding boxes. In the original Fast R-CNN \cite{dollar2014fast} framework, $crit$ is chosen as an smooth-L1 distance of coordinates in log-scale, while in \cite{yu2016unitbox}, $crit$ is chosen as the $- \ln(\text{IoU})$ between two bounding boxes.

Regression-based algorithms directly estimate the optimal solution $c^*$ with a feed-forward neural network. However, iterative bounding box regression methods are vulnerable to the change in the input distribution \cite{cai2017cascade} and may result in non-monotonic localization improvement, as shown in Figure~\ref{fig:apiter}.
To tackle these issues, we propose an optimization-based bounding box refinement method utilizing IoU-Net as a robust localization accuracy (IoU) estimator. Furthermore, IoU estimator can be used as an early-stop condition to implement iterative refinement with adaptive steps.
% and empirically provides a monotonic improvement in localization accuracy.

IoU-Net directly estimates $\mathrm{IoU}(box_{\text{det}}, box_{\text{gt}})$. While the proposed Precise RoI Pooling layer enables the computation of the gradient of IoU \wrt bounding box coordinates\footnote{We prefer Precise RoI-Pooling layer to RoI-Align layer \cite{he2017mask} as Precise RoI-Pooling layer is continuously differentiable \wrt the coordinates while RoI-Align is not.}, we can directly use gradient ascent method to find the optimal solution to Equation~\ref{eq:optimization}. Shown in Algorithm~\ref{alg:bbrefine}, viewing the estimation of the IoU as an optimization objective, we iteratively refine the bounding box coordinates with the computed gradient and maximize the IoU between the detected bounding box and its matched ground-truth. Besides, the predicted IoU is an interpretable indicator of the localization confidence on each bounding box and helps explain the performed transformation.

In the implementation, shown in Algorithm~\ref{alg:bbrefine} Line 6, we manually scale up the gradient \wrt the coordinates with the size of the bounding box on that axis (\eg, we scale up $\nabla x_1$~with $width(b_j)$). This is equivalent to perform the optimization in log-scaled coordinates ($x/w, y/h, \log w, \log h$) as in \cite{dollar2014fast}. We also employ a one-step bounding box regression for an initialization of the coordinates.

\begin{figure}[t]
       \centering
    \includegraphics[width=1\columnwidth]{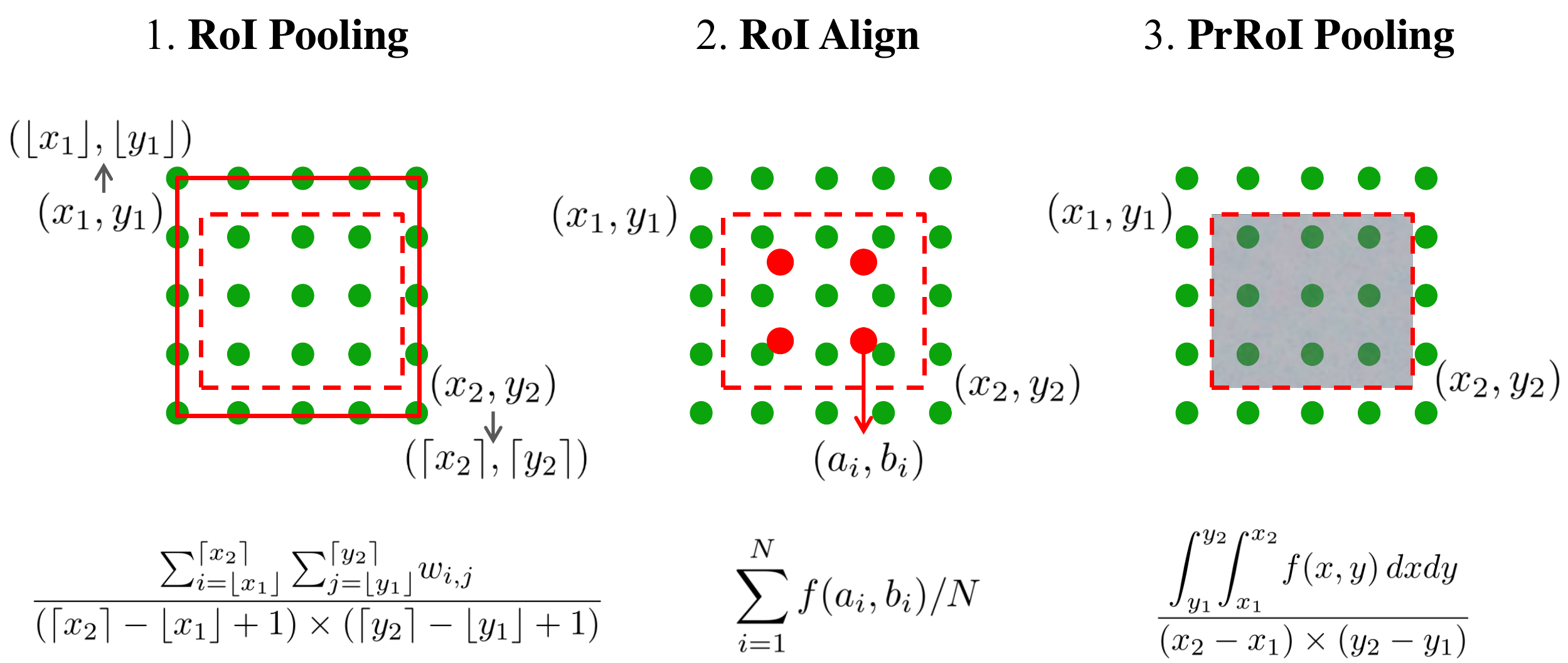}
    \caption{Illustration of RoI Pooling, RoI Align and PrRoI Pooling.}
    \label{fig:prroi}
    \vspace{-0.5em}
\end{figure}

\vspace{1em}
\myparagraph{Precise RoI Pooling.}
\newcommand{\prpool}{\mathrm{PrPool}}
We introduce Precise RoI Pooling (PrRoI Pooling, for short) powering our bounding box refinement\footnote{The code is released at: \url{https://github.com/vacancy/PreciseRoIPooling}}. It avoids any quantization of coordinates and has a continuous gradient on bounding box coordinates. Given the feature map $\mathcal{F}$ before RoI/PrRoI Pooling (\eg from Conv4 in ResNet-50), let $w_{i,j}$ be the feature at one discrete location $(i,j)$ on the feature map. Using bilinear interpolation, the discrete feature map can be considered continuous at any continuous coordinates $(x,y)$:
\begin{equation}
f(x,y) = \sum_{i,j}IC(x,y,i,j) \times w_{i,j},
\end{equation}
where $IC(x,y,i,j) = max(0,1-|x-i|)\times max(0,1-|y-j|)$ is the interpolation coefficient. Then denote a bin of a RoI as $bin=\{(x_1,y_1),(x_2,y_2)\}$, where $(x_1,y_1)$ and $(x_2,y_2)$ are the continuous coordinates of the top-left and bottom-right points, respectively. We perform pooling ({\it e.g.}, average pooling) given $bin$ and feature map $\mathcal{F}$ by computing a two-order integral:

\begin{equation}
\prpool(bin,\mathcal{F}) = \frac{ \displaystyle\int_{y1}^{y2}\hspace{-0.2cm}\int_{x1}^{x2}\!f(x,y)\,dxdy }{(x_2-x_1) \times (y_2-y_1)}.
\label{eqn:PrRoI}
\end{equation}

For a better understanding, we visualize RoI Pooling, RoI Align~\cite{he2017mask} and our PrRoI Pooing in Figure~\ref{fig:prroi}: in the traditional RoI Pooling, the continuous coordinates need to be quantized first to calculate the sum of the activations in the bin; to eliminate the quantization error, in RoI Align, $N=4$ continuous points are sampled in the bin, denoted as $(a_i, b_i)$, and the pooling is performed over the sampled points. Contrary to RoI Align where $N$ is pre-defined and not adaptive \wrt the size of the bin, the proposed PrRoI pooling directly compute the two-order integral based on the continuous feature map.

Moreover, based on the formulation in Equation~\ref{eqn:PrRoI},  $\prpool(Bin,\mathcal{F})$ is differentiable \wrt the coordinates of $bin$. For example, the partial derivative of $\prpool(B,\mathcal{F})$ \wrt $x_1$  could be computed as:
\begin{align}
\frac{\partial \prpool(bin,\mathcal{F})}{\partial x_1} &= \ \ \ \frac{\prpool(bin,\mathcal{F})}{x_2-x_1} - \frac{\int_{y1}^{y2}f(x_1,y)\,dy}{(x_2-x_1) \times (y_2-y_1)}.
\label{eqn:backpro}
\end{align}
The partial derivative of $\prpool(bin,\mathcal{F})$ \wrt other coordinates can be computed in the same manner. Since we avoids any quantization, $\prpool$ is continuously differentiable.
% which enables the bounding box refinement via the back-propagation through the PrRoI Pooling layer.

%% file: src/joint.tex
\subsection{Joint training}
\label{sec:joint}
The IoU predictor can be integrated into standard FPN pipelines for end-to-end training and inference. For clarity, we denote \emph{backbone} as the CNN architecture for image feature extraction and \emph{head} as the modules applied to individual RoIs.

Shown in Figure~\ref{fig:network}, the IoU-Net uses ResNet-FPN \cite{Lin_2017_CVPR} as the backbone, which has a top-down architecture to build a feature pyramid. FPN extracts features of RoIs from different levels of the feature pyramid according to their scale. The original RoI Pooling layer is replaced by the Precise RoI Pooling layer. As for the network head, the IoU predictor works in parallel with the R-CNN branch (including classification and bounding box regression) based on the same visual feature from the backbone.
% This IoU predictor introduces moderate computation overhead to the full pipeline as analyzed in Table~\ref{tab:exp:speed}. Moreover, this new branch and its training objective help the CNN learn more discriminative visual features, which improves the final detection performance.

We initialize weights from pre-trained ResNet models on ImageNet \cite{ILSVRC15}. All new layers are initialized with a zero-mean Gaussian with standard deviation 0.01 or 0.001. We use smooth-L1 loss for training the IoU predictor. The training data for the IoU predictor is separately generated as described in Section~\ref{sec:predictiou} within images in a training batch. IoU labels are normalized s.t. the values are distributed over $[-1, 1]$.

Input images are resized to have 800px along the short axis and a maximum of 1200px along the long axis. The classification and regression branch take 512 RoIs per image from RPNs. We use a batch size 16 for the training. The network is optimized for 160k iterations, with a learning rate of 0.01 which is decreased by a factor of 10 after 120k iterations. We also warm up the training by setting the learning rate to 0.004 for the first 10k iteration. We use a weight decay of 1e-4 and a momentum of 0.9.

During inference, we first apply bounding box regression for the initial coordinates. To speed up the inference, we first apply IoU-guided NMS on all detected bounding boxes. 100 bounding boxes with highest classification confidence are further refined using the optimization-based algorithm. 
We set $\lambda = 0.5$ as the step size, $\Omega_1 = 0.001$ as the early-stop threshold, $\Omega_2 = -0.01$ as the localization degeneration tolerance and $T = 5$ as the number of iterations.

%% file: src/experiments.tex
\section{Experiments}
We perform experiments on the 80-category MS-COCO detection dataset \cite{lin2014microsoft}. Following \cite{bell2016inside,Lin_2017_CVPR}, the models are trained on the union of 80k training images and 35k validation images (\emph{trainval35k}) and evaluated on a set of 5k validation images (\emph{minival}). To validate the proposed methods, in both Section \ref{sec:expr:nms} and \ref{sec:expr:refine}, a \emph{standalone} IoU-Net (without R-CNN modules) is trained separately with the object detectors. IoU-guided NMS and optimization-based bounding box refinement, powered by the IoU-Net, are applied to the detection results.

\subsection{IoU-guided NMS}
\label{sec:expr:nms}
Table~\ref{tab:exp:iou} summarizes the performance of different NMS methods. While Soft-NMS preserve more bounding boxes (there is no real ``suppression''), IoU-guided NMS improves the results by improving the localization of the detected bounding boxes. As a result, IoU-guided NMS performs significantly better than the baselines on high IoU metrics (\eg, AP${}_{\text{90}}$).

We delve deeper into the behavior of different NMS algorithms by analyzing their recalls at different IoU threshold. The raw detected bounding boxes are generated by a ResNet50-FPN without any NMS. As the requirement of localization accuracy increases, the performance gap between IoU-guided NMS and other methods goes larger. In particular, the recall at matching IoU $\mathrm{\Omega_{test}}=0.9$ drops to 18.7\% after traditional NMS, while the IoU-NMS reaches 28.9\% and the No-NMS ``upper bound'' is 39.7\%.

\begin{table}[!h]
\begin{center}
\setlength{\tabcolsep}{0.7mm}{
\begin{tabular}{l|cc|c|ccccc}
\hline\hline
Method &~~+Soft-NMS~~&~~+IoU-NMS~~&~~$\mathrm{AP}$~~~&    $\mathrm{AP_{50}}$    &    $\mathrm{AP_{60}}$    &    $\mathrm{AP_{70}}$    &    $\mathrm{AP_{80}}$    &    $\mathrm{AP_{90}}$    \\
\hline
\multirow{3}{*}{FPN}
&        &                & 36.4 & \textbf{58.0} & \textbf{53.1} & 44.9 & 31.2 & 9.8 \\    
&$\checkmark$&            & 36.8 & 57.5 & \textbf{53.1} & \textbf{45.7} & 32.3 & 10.3 \\    
&        &$\checkmark$    & \textbf{37.3} & 56.0 & 52.2 & 45.6 & \textbf{33.9} & \textbf{13.3} \\
\hline
\multirow{3}{*}{Cascade R-CNN}
&        &             & 40.6 & \textbf{59.3} & 55.2 & 49.1 & 38.7 & 16.7 \\
&$\checkmark$&         & \textbf{40.9} & 58.2 & \textbf{54.7} & \textbf{49.4} & \textbf{39.9} & 17.8 \\
&        &$\checkmark$& 40.7 & 58.0 & \textbf{54.7} & 49.2 & 38.8 & \textbf{18.9} \\
\hline
\multirow{3}{*}{Mask-RCNN}
&        &             & 37.5 & \textbf{58.6} & \textbf{53.9} & 46.3 & 33.2 & 10.9 \\
&$\checkmark$&         & 37.9 & 58.2 & \textbf{53.9} & \textbf{47.1} & 34.4 & 11.5 \\
&        &$\checkmark$& \textbf{38.1} & 56.4 & 52.7 & 46.7 & \textbf{35.1} & \textbf{14.6} \\
\hline
\end{tabular}
}
\end{center}
\caption{Comparison of IoU-guided NMS with other NMS methods. By preserving bounding boxes with accurate localization, IoU-guided NMS shows significant improvement in AP with high matching IoU threshold (\eg, AP${}_{\text{90}}$).
% IoU-guided NMS provides no extra overhead except the feed-forwarding of the IoU predictor.
}
\label{tab:exp:iou}
\vspace{-2em}
\end{table}

\begin{SCfigure}
\centering
\caption{
Recall curves of different NMS methods at different IoU threshold for matching detected bounding boxes with the ground-truth. No-NMS (no box is suppressed) is provided as the upper bound of the recall.
The proposed IoU-NMS has a higher recall and effectively narrows the gap to the upper-bound at high IoU threshold (\eg, 0.8).
}
\label{fig:recall}
\includegraphics[width=0.55\linewidth]{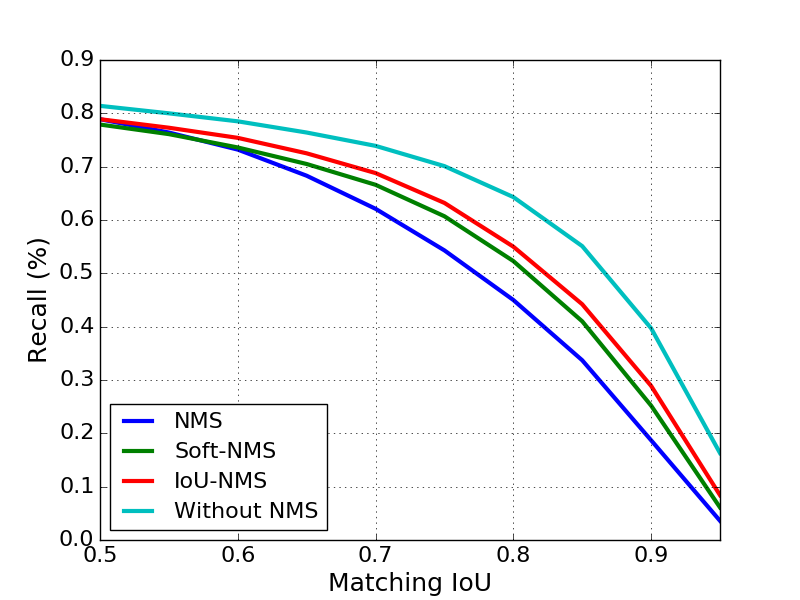}
\end{SCfigure}

\subsection{Optimization-based bounding box refinement}
\label{sec:expr:refine}

\begin{table}[!t]
\centering
\setlength{\tabcolsep}{2.2mm}{
  \begin{tabular*}{\linewidth}{l|c|c|ccccc}
  \hline \hline
    Method & +Refinement & $\mathrm{AP}$ & $\mathrm{AP_{50}}$& $\mathrm{AP_{60}}$    & $\mathrm{AP_{70}}$    & $\mathrm{AP_{80}}$    & $\mathrm{AP_{90}}$\\
  \hline
      \multirow{2}{*}{FPN} 
            &              & 36.4 & \textbf{58.0} & \textbf{53.1} & 44.9 & 31.2 & 9.8 \\
             &$\checkmark$& \textbf{38.0} & 57.7 & \textbf{53.1} & \textbf{46.1} & \textbf{34.3} & \textbf{14.6} \\
  \hline
      \multirow{2}{*}{Cascade R-CNN}
            &              & 40.6 & \textbf{59.3} & 55.2 & 49.1 & 38.7 & 16.7 \\
            &$\checkmark$& \textbf{41.4} & \textbf{59.3} & \textbf{55.3} & \textbf{49.6} & \textbf{39.4} & \textbf{19.5} \\
  \hline
      \multirow{2}{*}{Mask-RCNN}
        &             & 37.5 & \textbf{58.6} & \textbf{53.9} & 46.3 & 33.2 & 10.9 \\
        &$\checkmark$& \textbf{39.2} & 57.9 & 53.6 & \textbf{47.4} & \textbf{36.5} & \textbf{16.4}\\
  \hline
  \end{tabular*}}
\caption{The optimization-based bounding box refinement further improves the performance of several CNN-based object detectors.}
\label{tab:exp:iourefine}
\end{table}

The proposed optimization-based bounding box refinement is compatible with most of the CNN-based object detectors \cite{Lin_2017_CVPR,cai2017cascade,he2017mask}, as shown in Table~\ref{tab:exp:iourefine}. Applying the bounding box refinement after the original pipelines with the \emph{standalone} IoU-Net further improve the performance by localizing object more accurately. The refinement further improves $\text{AP}_\text{90}$ by $2.8\%$ and the overall AP by $0.8\%$ even for Cascade R-CNN which has a three-stage bounding box regressor.

\subsection{Joint training}
%Independent IoU network is very fast to train. Training with ResNet-50-FPN on COCO trainval35k takes 24 hours in our synchronized 8-GPU implementation ( 0.55s per 16 mini-batch). As for the IoU network with classification and bounding box branch, it takes 36 hours to train on the same settings. 

IoU-Net can be end-to-end optimized in parallel with object detection frameworks. We find that adding IoU predictor to the network helps the network to learn more discriminative features which improves the overall AP by 0.6 and 0.4 percent for ResNet50-FPN and ResNet101-FPN respectively. The IoU-guided NMS and bounding box refinement further push the performance forward. We achieve $40.6\%$ AP with ResNet101-FPN compared to the baseline $38.5\%$ (improved by $2.1\%$). The inference speed is demonstrated in Table~\ref{tab:exp:joint}, showing that IoU-Net improves the detection performance with tolerable computation overhead.

We mainly attribute the inferior results on AP${}_{50}$ in Table~\ref{tab:exp:joint} to the IoU estimation error. When the bounding boxes have a lower IoU with the ground-truth, they have a larger variance in appearance. Visualized in Figure~\ref{fig:distribution:iou}, the IoU estimation becomes less accurate for boxes with lower IoU. This degenerates the performance of the downstream refinement and suppression. We empirically find that this problem can be partially solved by techniques such as sampling more bounding boxes with lower IoU during the training.

\begin{table}[!t]
\centering
\setlength{\tabcolsep}{1mm}{
  \begin{tabular}{l|c|cc|cccccc}
  \hline \hline
    Backbone & Method & +IoU-NMS~&~+Refine &~~$\mathrm{AP}$~~~& $\mathrm{AP_{50}}$ & $\mathrm{AP_{60}}$ & $\mathrm{AP_{70}}$ & $\mathrm{AP_{80}}$ & $\mathrm{AP_{90}}$\\
    \hline
    \multirow{4}{*}{ResNet-50} &FPN& &         & 36.4 & 58.0 & 53.1 & 44.9 & 31.2 & 9.8 \\
    \cline{2-2} \cline{5-10}
    &\multirow{3}{*}{IoU-Net} & &         & 37.0 & \textbf{58.3} & \textbf{53.8} & 45.7 & 31.9 & 10.7 \\    
    & &$\checkmark$&                         & 37.6 & 56.2 & 52.4 & 46.0 & 34.1 & 14.0 \\    
    & &$\checkmark$&$\checkmark$            & \textbf{38.1} & 56.3 & 52.4 & \textbf{46.3} & \textbf{35.1} & \textbf{15.5} \\
    \hline
    \multirow{4}{*}{ResNet-101} &FPN& &     & 38.5 & \textbf{60.3} & \textbf{55.5} & 47.6 & 33.8 & 11.3 \\
    \cline{2-2} \cline{5-10}
    &\multirow{3}{*}{IoU-Net} & &         & 38.9 & 60.2 & \textbf{55.5} & 47.8 & 34.6 & 12.0 \\
    & &$\checkmark$&                        & 40.0 & 59.0 & 55.1 & 48.6 & 37.0 & 15.5 \\    
    & &$\checkmark$&$\checkmark$            & \textbf{40.6} & 59.0 & 55.2 & \textbf{49.0} & \textbf{38.0} & \textbf{17.1} \\\hline
  \end{tabular}
}
\caption{Final experiment results on MS-COCO. IoU-Net denotes ResNet-FPN embedded with IoU predictor. We improve the FPN baseline by $\approx 2\%$ in AP.}
\label{tab:exp:joint}
\end{table}

\begin{table}[!t]
\centering
\setlength{\tabcolsep}{1.3mm}{
  \begin{tabular}{l|c|c|c|c}
      \hline\hline
    Method &~~~~FPN~~~~~&~Mask-RCNN~~& Cascade R-CNN &~~~~IoU-Net~~~~~\\
    \hline
    ~Speed (sec./image)~  &0.255 & 0.267     & 0.384        & 0.305 \\
    \hline
  \end{tabular}
}
\caption{Inference speed of multiple object detectors on a single TITAN X GPU. The models share the same backbone network ResNet50-FPN. The input resolution is 1200x800. All hyper-parameters are set to be the same.}
\label{tab:exp:speed}
\end{table}

%% file: samplepaper.bbl
\begin{thebibliography}{10}
\providecommand{\url}[1]{\texttt{#1}}
\providecommand{\urlprefix}{URL }
\providecommand{\doi}[1]{https://doi.org/#1}

\bibitem{bell2016inside}
Bell, S., Lawrence~Zitnick, C., Bala, K., Girshick, R.: Inside-outside net:
  Detecting objects in context with skip pooling and recurrent neural networks.
  In: Proceedings of the IEEE Conference on Computer Vision and Pattern
  Recognition. pp. 2874--2883 (2016)

\bibitem{bodla2017improving}
Bodla, N., Singh, B., Chellappa, R., Davis, L.S.: Improving object detection
  with one line of code. arXiv preprint arXiv:1704.04503  (2017)

\bibitem{cai2017cascade}
Cai, Z., Vasconcelos, N.: Cascade r-cnn: Delving into high quality object
  detection. arXiv preprint arXiv:1712.00726  (2017)

\bibitem{dalal2005histograms}
Dalal, N., Triggs, B.: Histograms of oriented gradients for human detection.
  In: Computer Vision and Pattern Recognition, 2005. CVPR 2005. IEEE Computer
  Society Conference on. vol.~1, pp. 886--893. IEEE (2005)

\bibitem{dollar2014fast}
Doll{\'a}r, P., Appel, R., Belongie, S., Perona, P.: Fast feature pyramids for
  object detection. IEEE Transactions on Pattern Analysis and Machine
  Intelligence  \textbf{36}(8),  1532--1545 (2014)

\bibitem{gidaris2015object}
Gidaris, S., Komodakis, N.: Object detection via a multi-region and semantic
  segmentation-aware cnn model. In: Proceedings of the IEEE International
  Conference on Computer Vision. pp. 1134--1142 (2015)

\bibitem{gidaris2016attend}
Gidaris, S., Komodakis, N.: Attend refine repeat: Active box proposal
  generation via in-out localization. arXiv preprint arXiv:1606.04446  (2016)

\bibitem{Girshick_2015_ICCV}
Girshick, R.: Fast r-cnn. In: The IEEE International Conference on Computer
  Vision (ICCV) (December 2015)

\bibitem{Girshick_2014_CVPR}
Girshick, R., Donahue, J., Darrell, T., Malik, J.: Rich feature hierarchies for
  accurate object detection and semantic segmentation. In: The IEEE Conference
  on Computer Vision and Pattern Recognition (CVPR) (June 2014)

\bibitem{he2017mask}
He, K., Gkioxari, G., Doll{\'a}r, P., Girshick, R.: Mask r-cnn. In: The IEEE
  International Conference on Computer Vision (ICCV) (2017)

\bibitem{hosang2017learning}
Hosang, J., Benenson, R., Schiele, B.: Learning non-maximum suppression. arXiv
  preprint  (2017)

\bibitem{hu2017relation}
Hu, H., Gu, J., Zhang, Z., Dai, J., Wei, Y.: Relation networks for object
  detection. arXiv preprint arXiv:1711.11575  (2017)

\bibitem{hu2017fastmask}
Hu, H., Lan, S., Jiang, Y., Cao, Z., Sha, F.: Fastmask: Segment multi-scale
  object candidates in one shot. In: Proceedings of the IEEE Conference on
  Computer Vision and Pattern Recognition. pp. 991--999 (2017)

\bibitem{huang2015densebox}
Huang, L., Yang, Y., Deng, Y., Yu, Y.: Densebox: Unifying landmark localization
  with end to end object detection. arXiv preprint arXiv:1509.04874  (2015)

\bibitem{li2017multi}
Li, J., Liang, X., Li, J., Wei, Y., Xu, T., Feng, J., Yan, S.: Multi-stage
  object detection with group recursive learning. IEEE Transactions on
  Multimedia  (2017)

\bibitem{Lin_2017_CVPR}
Lin, T.Y., Doll{\'a}r, P., Girshick, R., He, K., Hariharan, B., Belongie, S.:
  Feature pyramid networks for object detection. In: The IEEE Conference on
  Computer Vision and Pattern Recognition (CVPR) (2017)

\bibitem{lin2014microsoft}
Lin, T.Y., Maire, M., Belongie, S., Hays, J., Perona, P., Ramanan, D.,
  Doll{\'a}r, P., Zitnick, C.L.: Microsoft coco: Common objects in context. In:
  European conference on computer vision. pp. 740--755. Springer (2014)

\bibitem{najibi2016g}
Najibi, M., Rastegari, M., Davis, L.S.: G-cnn: an iterative grid based object
  detector. In: Proceedings of the IEEE Conference on Computer Vision and
  Pattern Recognition. pp. 2369--2377 (2016)

\bibitem{pinheiro2015learning}
Pinheiro, P.O., Collobert, R., Doll{\'a}r, P.: Learning to segment object
  candidates. In: Advances in Neural Information Processing Systems. pp.
  1990--1998 (2015)

\bibitem{pinheiro2016learning}
Pinheiro, P.O., Lin, T.Y., Collobert, R., Doll{\'a}r, P.: Learning to refine
  object segments. In: European Conference on Computer Vision. pp. 75--91.
  Springer (2016)

\bibitem{rajaram2016refinenet}
Rajaram, R.N., Ohn-Bar, E., Trivedi, M.M.: Refinenet: Iterative refinement for
  accurate object localization. In: Intelligent Transportation Systems (ITSC),
  2016 IEEE 19th International Conference on. pp. 1528--1533. IEEE (2016)

\bibitem{ren2015faster}
Ren, S., He, K., Girshick, R., Sun, J.: Faster r-cnn: Towards real-time object
  detection with region proposal networks. In: Advances in neural information
  processing systems. pp. 91--99 (2015)

\bibitem{NIPS2015_5638}
Ren, S., He, K., Girshick, R., Sun, J.: Faster r-cnn: Towards real-time object
  detection with region proposal networks. In: Cortes, C., Lawrence, N.D., Lee,
  D.D., Sugiyama, M., Garnett, R. (eds.) Advances in Neural Information
  Processing Systems 28, pp. 91--99. Curran Associates, Inc. (2015),
  \url{http://papers.nips.cc/paper/5638-faster-r-cnn-towards-real-time-object-detection-with-region-proposal-networks.pdf}

\bibitem{rothe2014non}
Rothe, R., Guillaumin, M., Van~Gool, L.: Non-maximum suppression for object
  detection by passing messages between windows. In: Asian Conference on
  Computer Vision. pp. 290--306. Springer (2014)

\bibitem{ILSVRC15}
Russakovsky, O., Deng, J., Su, H., Krause, J., Satheesh, S., Ma, S., Huang, Z.,
  Karpathy, A., Khosla, A., Bernstein, M., Berg, A.C., Fei-Fei, L.: {ImageNet
  Large Scale Visual Recognition Challenge}. International Journal of Computer
  Vision (IJCV)  \textbf{115}(3),  211--252 (2015).
  \doi{10.1007/s11263-015-0816-y}

\bibitem{taigman2014deepface}
Taigman, Y., Yang, M., Ranzato, M., Wolf, L.: Deepface: Closing the gap to
  human-level performance in face verification. In: Proceedings of the IEEE
  conference on computer vision and pattern recognition. pp. 1701--1708 (2014)

\bibitem{toshev2014deeppose}
Toshev, A., Szegedy, C.: Deeppose: Human pose estimation via deep neural
  networks. In: Proceedings of the IEEE conference on computer vision and
  pattern recognition. pp. 1653--1660 (2014)

\bibitem{uijlings2013selective}
Uijlings, J.R., Van De~Sande, K.E., Gevers, T., Smeulders, A.W.: Selective
  search for object recognition. International journal of computer vision
  \textbf{104}(2),  154--171 (2013)

\bibitem{wang2017repulsion}
Wang, X., Xiao, T., Jiang, Y., Shao, S., Sun, J., Shen, C.: Repulsion loss:
  Detecting pedestrians in a crowd. arXiv preprint arXiv:1711.07752  (2017)

\bibitem{wu2017learning}
Wu, J., Lu, E., Kohli, P., Freeman, W.T., Tenenbaum, J.B.: Learning to see
  physics via visual de-animation. In: Advances in Neural Information
  Processing Systems (2017)

\bibitem{yang2016craft}
Yang, B., Yan, J., Lei, Z., Li, S.Z.: Craft objects from images. arXiv preprint
  arXiv:1604.03239  (2016)

\bibitem{yu2016unitbox}
Yu, J., Jiang, Y., Wang, Z., Cao, Z., Huang, T.: Unitbox: An advanced object
  detection network. In: Proceedings of the 2016 ACM on Multimedia Conference.
  pp. 516--520. ACM (2016)

\bibitem{zitnick2014edge}
Zitnick, C.L., Doll{\'a}r, P.: Edge boxes: Locating object proposals from
  edges. In: European Conference on Computer Vision. pp. 391--405. Springer
  (2014)

\end{thebibliography}
